\documentclass[letterpaper]{article} 
\usepackage{aaai2026}  
\usepackage{times}  
\usepackage{helvet}  
\usepackage{courier}  
\usepackage[hyphens]{url}  
\usepackage{graphicx} 
\usepackage{natbib}  
\usepackage{caption} 
\usepackage{algorithm}
\usepackage{algorithmic}
\usepackage{comment}

\usepackage{amssymb}
\usepackage{amsmath}
\usepackage{booktabs}

\usepackage{multirow}
\usepackage{times}
\usepackage{soul}
\usepackage{url}
\usepackage{caption}

\usepackage[switch]{lineno}
\usepackage{color}
\usepackage{comment}
\usepackage{float}
\usepackage{enumitem}
\usepackage{newfloat}
\usepackage{listings}
\DeclareCaptionStyle{ruled}{labelfont=normalfont,labelsep=colon,strut=off} 
\floatstyle{ruled}
\newfloat{listing}{tb}{lst}{}
\floatname{listing}{Listing}
\title{DP-GPT4MTS: Dual-Prompt Large Language Model for Textual-Numerical Time Series Forecasting}
\author{
    Chanjuan Liu\textsuperscript{\rm 1}, Shengzhi Wang\textsuperscript{\rm 2}, Enqiang Zhu\textsuperscript{\rm 2}\thanks{Corresponding author: zhuenqiang@gzhu.edu.cn}\\
}
\affiliations{
    \textsuperscript{\rm 1}School of Computer Science and Technology,
Dalian University of Technology, Dalian 116024, China\\
    \textsuperscript{\rm 2} Institute of Computing Technology, Guangzhou
University, Guangzhou 510006, China\\
}

\usepackage{bibentry}

\begin{document}

\maketitle

\begin{abstract}
Time series forecasting is crucial in strategic planning and decision-making across various industries. Traditional forecasting models mainly concentrate on numerical time series data, often overlooking important textual information such as events and news, which can significantly affect forecasting accuracy. While large language models offer a promise for integrating multimodal data, existing single-prompt frameworks struggle to effectively capture the semantics of timestamped text, introducing redundant information that can hinder model performance. To address this limitation, we introduce DP-GPT4MTS (Dual-Prompt GPT2-base for Multimodal Time Series), a novel dual-prompt large language model framework that combines two complementary prompts: an explicit prompt for clear task instructions and a textual prompt for context-aware embeddings from time-stamped data.  The tokenizer generates the explicit prompt while the embeddings from the textual prompt are refined through self-attention and feed-forward networks. Comprehensive experiments conducted on diverse textural-numerical time series datasets demonstrate that this approach outperforms state-of-the-art algorithms in time series forecasting. This highlights the significance of incorporating textual context via a dual-prompt mechanism to achieve more accurate time series predictions\footnote{Code and Datasets are provided in the supplementary materials accompanying this paper.}.
\end{abstract}

\section{Introduction}
Time series forecasting is an essential method that leverages historical data to predict future trends and values, serving as a cornerstone for strategic planning and decision-making across various domains \cite{1}. For instance, in financial analysis, precise predictions for investment strategies, risk assessment, and market evaluations enable decision-makers to anticipate market fluctuations, evaluate investment opportunities, and mitigate risks \cite{2,3}. In supply chain management, accurate forecasts enhance inventory control, demand planning, and logistics optimization, ultimately improving operational efficiency and reducing costs \cite{4}. Moreover, effective traffic forecasting aids in route planning and vehicle scheduling, alleviating congestion and fostering better transportation systems \cite{5}. 


In today’s data-rich environment, time series data often includes textual information such as news articles and event reports, which we refer to as textual-numerical time series data. Integrating numerical time series with textual insights is essential for increasing forecasting accuracy. This fusion provides a more nuanced understanding of the factors driving trends. For example, in financial analysis, combining time series data (e.g., stock prices) with news reports allows models to capture the causal relationships behind market fluctuations better. Similarly, in traffic flow forecasting, incorporating text information like weather forecasts, road construction announcements, or special event details helps models understand the causes of traffic changes more accurately. This forecasting method, which we call textual-numerical time series forecasting, combines numerical data with relevant text. This approach is essential for making more accurate and informed predictions, as it mirrors how people typically integrate both types of information when making decisions in the real world.

Current time series forecasting methods, ranging from traditional techniques such as Autoregressive Integrated Moving Average (ARIMA) \cite{6}, exponential smoothing \cite{7}, and spectral analysis \cite{7} to machine learning approaches like Transformer models \cite{26,38} and linear models \cite{9}, have shown remarkable effectiveness in their respective applications. However, many of these methods primarily focus on numerical sequences, often overlooking the valuable contextual information provided by textual data \cite{10}.

\begin{figure*}[htbp] 
    \centering
    \includegraphics[width=0.9\textwidth]{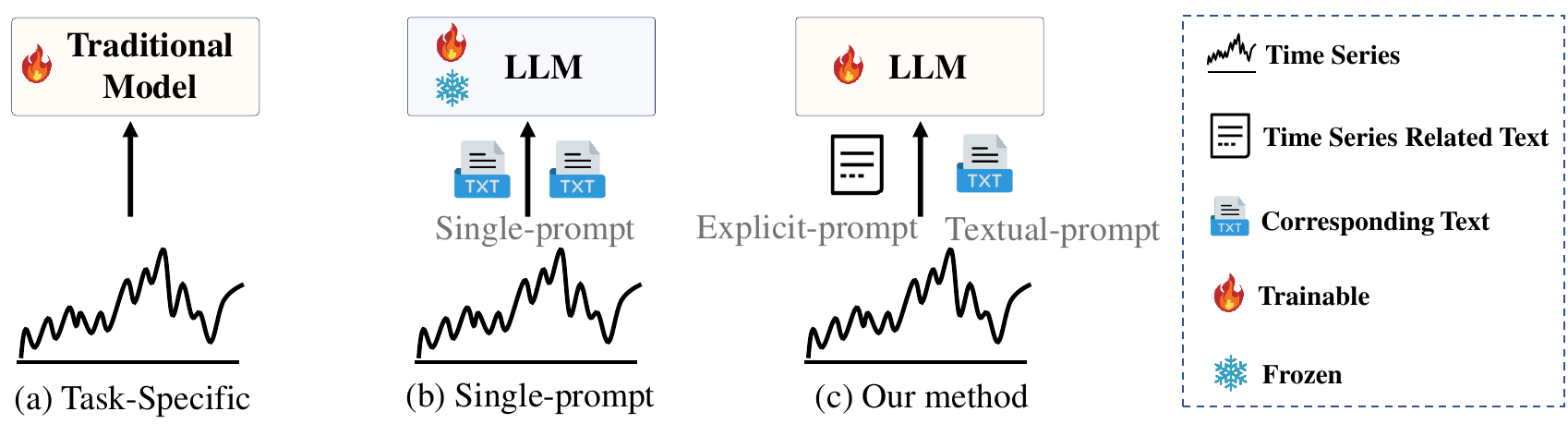} 
    \caption{Illustration of the main difference between DP-GPT4MTS and other methods.} 
    \label{fig:1} 
\end{figure*}

In recent years, pre-trained foundational models, especially large language models (LLMs) \cite{11}, have demonstrated remarkable capabilities in time series forecasting \cite{12,13,14}. Several studies have started to interpret time series data as text sequences \cite{17,18,19}, while others have focused on integrating time series data into embeddings for input into large models \cite{15,16,10}. Notably, TimeLLM \cite{16} employs prompt-as-prefix and reprogramming techniques to align text prototypes. In contrast, GPT4MTS \cite{10} uses textual information as separate soft prompts, combined with time series data for input into large models. This dual approach aims to capture temporal patterns and contextual insights more effectively. However, most existing LLMs rely on a single prompt tailored for specific input data or task types. This method presents significant challenges when dealing with multimodal time series data incorporating additional textual information. It often leads to redundant information and a lack of precision in capturing the relevance and significance of the text, thereby limiting the model's ability to fully leverage the rich contextual insights available.

\vspace{0.1cm}
\noindent\textbf{Our contribution}  
To address the limitations in textual-numerical time series forecasting, we present DP-GPT4MTS (Dual-Prompt GPT2-base for Multimodal Time Series), a novel dual-prompt large language model framework. The main distinction between the proposed framework and existing methods for time series forecasting is illustrated in Figure \ref{fig:1}. Unlike the traditional models for time series forecasting, which are task-specific, and other LLMs that rely on a single prompt, DP-GPT4MTS employs two complementary prompt mechanisms. The first is an explicit prompt, which acts as a fixed prefix providing clear task instructions and human-readable guidance to enhance the model's predictive capabilities. The second is a textual prompt that processes time-stamped textual data to generate embeddings, allowing for contextual adaptation during training. The tokenizer of the pre-trained large model generates embeddings for the explicit prompt,  which helps the model better grasp the semantic information conveyed by the natural language description. The embeddings generated via the textural prompt are refined through self-attention mechanisms and feed-forward networks. By integrating these two prompts, our model fully combines textual and numerical information, thereby improving prediction accuracy for complex textual-numerical time series tasks. 

Comprehensive experiments on various textual-numerical time series datasets show that DP-GPT4MTS exceeds current methods in prediction accuracy. This framework is, to our knowledge, the first dual-prompt approach specifically designed for forecasting textual-numerical time series.

\vspace{0.1cm}

\vspace{0.1cm}
\noindent\textbf{Paper organization} The rest of this paper is structured as follows. Section \ref{sec-2} gives a brief review of the related work. Section \ref{sec-3} describes our DP-GPT4MTS framework in detail, covering the problem statement, model structure, and key components. Section \ref{sec-4} presents the experimental results and compares our method with other state-of-the-art techniques. Lastly, Section \ref{sec-5} summarizes the main contributions of this paper and suggests areas for future research.

\section{Related Work} \label{sec-2}
\textbf{Time series forecasting} Traditional time series forecasting methods involve analyzing historical data and employing statistical models to predict future trends based on the assumption that past patterns will persist \cite{21,22}. However, these approaches often face limitations when dealing with large-scale datasets \cite{23}. The emergence of deep learning has introduced a variety of time series forecasting networks  \cite{9,20,25,26,39} that excel at capturing nonlinear relationships and dependencies directly from historical data, making them particularly effective for managing larger and more complex datasets. Nonetheless, these deep learning methods predominantly focus on the numerical aspects of time series data and are not equipped to directly process the associated textual information.

\vspace{0.1cm}
\noindent\textbf{LLMs for time series forecasting}  LLMs, including the GPT series \cite{8,11,27} and LLaMa \cite{28}, have exhibited outstanding performance across a range of natural language processing (NLP) tasks. With their extensive parameter sets, these models acquire a wide array of general knowledge and reasoning skills during the pretraining phase, essential for developing intelligent systems capable of commonsense reasoning. As the demand for foundational models specifically designed for time series data continues, recent innovations such as ForecastPFN \cite{29} and TimeGPT \cite{30} mark significant strides in time series analysis. While these models effectively capture the unique temporal dynamics and patterns within the domain, their limitations in scale and variability have historically obstructed the development of general-purpose models. 

Adaptive LLMs for time series analysis has been proposed to address this challenge. This approach aims to harness their pre-trained capabilities to tackle a range of downstream tasks effectively, with a particular focus on enhancing effectiveness, efficiency, and interpretability. 
There are two primary adaptation paradigms: the embedding-visible paradigm \cite{10,15,16}, which integrates time series data into embeddings for input into large models, and the text-visible paradigm \cite{17,18,19}, which treats time series data as textual sequences for processing. The main distinction between these approaches lies in how the time series data are integrated and the methods used for input and output. Research indicates that models like FPT \cite{15} can successfully perform time series tasks even with frozen LLM parameters by harnessing the universality of self-attention mechanisms. However, existing methods often depend on single-prompt operations when addressing textual-numerical time series data. This reliance leads to difficulties in effectively extracting vital features from the accompanying textual information and introduces excessive redundancy, ultimately hindering the prediction performance of numerical embeddings.

\vspace{0.1cm}
\noindent\textbf{Prompt design} Prompt-based techniques involve transforming input text into specific templates and reorganizing tasks to leverage the capabilities of pre-trained language models fully \cite{32}. However, in the context of textual-numerical time series data, existing LLMs, such as GPT4MTS \cite{10} and TimeLLM \cite{16}, rely on single-prompt designs that are unable to effectively extract key information from textual data for guidance. Thus, exploring new prompting methods is essential. Our proposed framework consists of two complementary prompt mechanisms: An explicit prompt providing clear task instructions and a textual prompt enabling context-aware adaptation during training. Together, these mechanisms leverage the rich contextual insights inherent in numerical data, enhancing prediction accuracy for time series forecasting.


\section{The DP-GPT4MTS Framework} \label{sec-3}

This section introduces our DP-GPT4MTS framework. We begin by clearly defining the problem of time series forecasting using both textual and numerical data. Then, we present an overview of the proposed framework, focusing on its key components that are designed to improve forecasting performance effectively. 

\subsection{Problem Definition}
Let \( D = \{ ([x_1, s_1], \ldots, [x_n, s_n]) \} \) be a textual-numerical time series dataset, where  $x_t$ (for $1\leq t \leq n$) denotes the numerical value at timestamp \( t \), \( s_t \) represents the textual summary linked to timestamp \( t \), and  \( n \) indicates the total length of the time series dataset. Suppose that there is a series of univariate time series samples within a lookback window of length \( L \), accompanied by their corresponding textual summaries \( ([x_1, s_1], \dots, [x_L, s_L]) \). We intend to predict the values for the subsequent \( T \) timestamps, specifically \( [x_{L+1}], \ldots, [x_{L+T}] \). To accomplish this, we need to develop a mapping function \( f: \left( [x_1, s_1], \ldots, [x_L, s_L] \right) \to \left( [x_{L+1}], \ldots, [x_{L+T}] \right) \). This function will be designed to learn from the numerical time series data and the associated textual summaries, enabling it to forecast future values accurately.

\subsection{Model Overview}
\begin{figure}[t] 
    \centering
    \includegraphics[width=0.5\textwidth]{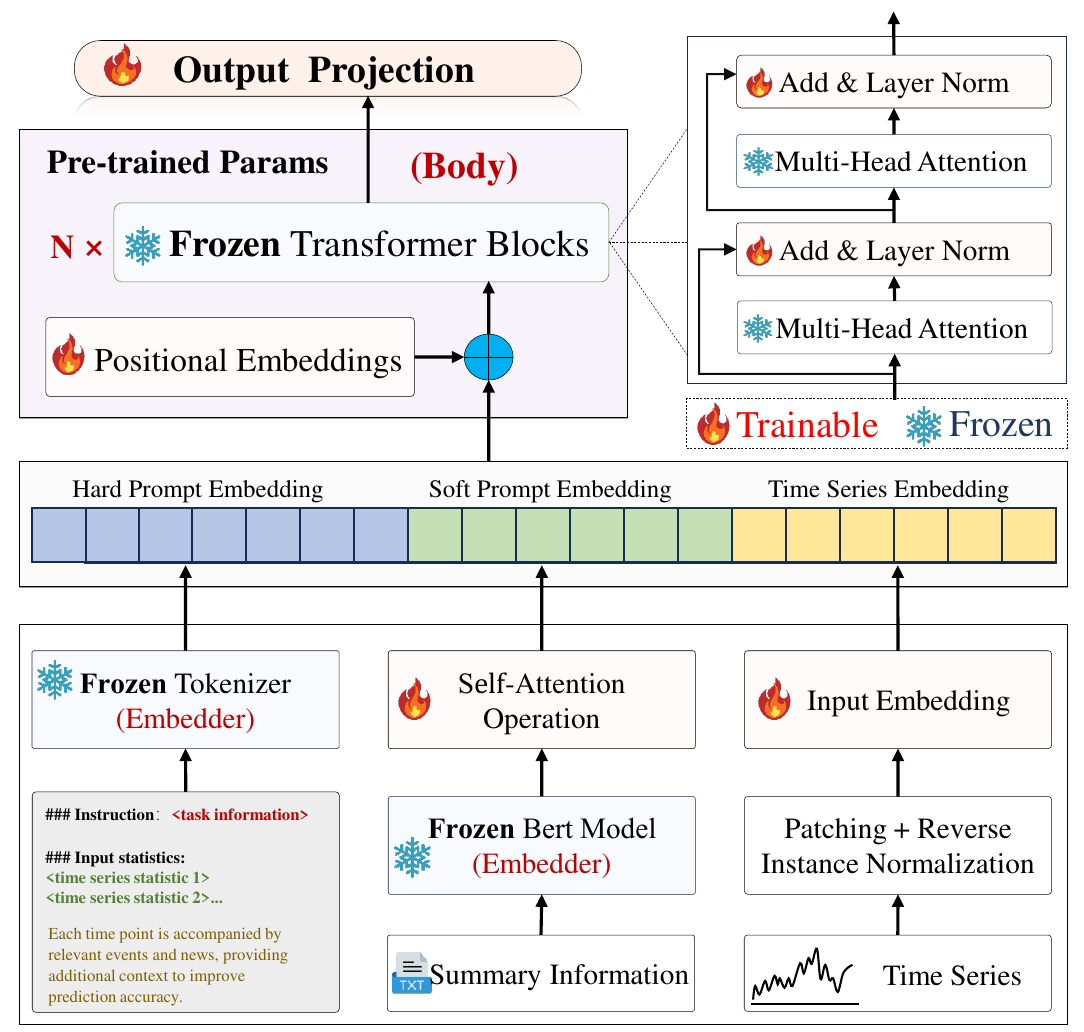} 
    \caption{Overview of our DP-GPT4MTS model.} 
    \label{fig:2} 
\end{figure}

As Figure \ref{fig:2} shows, DP-GPT4MTS leverages a dual-prompt structure. The process begins with an explicit prompt functioning as a hard prompt, which acts as an embedding prefix. Following this, the pre-trained language model BERT generates a textual summary, extracting classification (CLS) semantic information. Subsequently, a self-attention mechanism is employed to create soft prompt embeddings. Finally, the time series data undergoes patch slicing and reverse instance normalization, resulting in the formation of input embeddings. These embeddings, coupled with the dual prompts, constitute the input to the large model.

\subsection{Explicit Prompt-as-FirstPrefix}
\begin{figure}[hbtp] 
    \centering
    \includegraphics[width=0.45\textwidth]{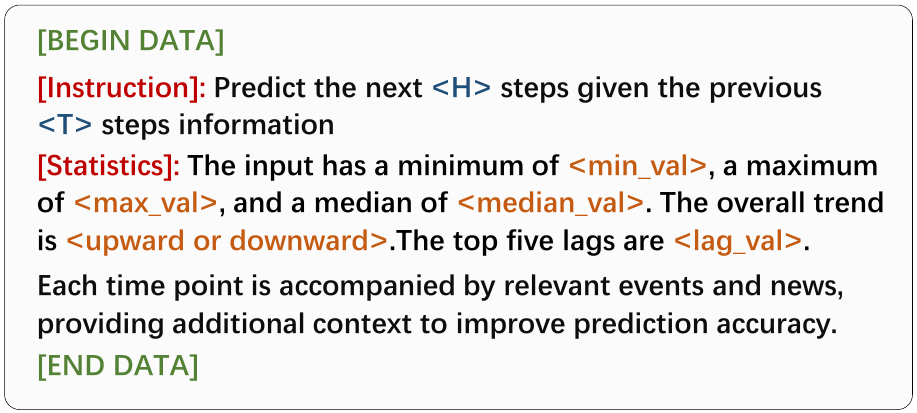} 
    \caption{Explicit prompt example. \texttt{<>} are task-specific configurations and calculated input statistics.} 
    \label{fig:3} 
\end{figure}

Inspired by TimeLLM \cite{16}, we adopt an explicit Prompt-as-First-Prefix strategy to guide the model's prediction. In this framework, we prepend a prompt embedding to the input sequence, which encodes rich contextual knowledge and domain priors. Our prompt consists of three core components: (1) task instructions, (2) input statistics, and (3) natural language explanations. At each time step, news and events contribute additional textual information. A simple illustration is provided in Figure \ref{fig:3}, where we suggest that this supplementary textual data can enhance the model's understanding of the contextual information within the dataset. Additionally, we incorporate significant statistical insights, such as trends and lags, to enrich the explicit prompt, thereby facilitating pattern recognition and reasoning. By employing the frozen tokenizer of the backbone language model, we generate the explicit prompt embedding \( E \in \mathbb{R}^{w \times D} \), where \( w \) denotes the number of tokens in the prompt and \( D \) denotes the hidden dimension of the backbone language model.

\subsection{Textual Prompt-as-SecondPrefix}
To process time series textual information of length \( L \), we employ a pre-trained language model, BERT, to extract the CLS semantic information and generate a semantic vector \( S \in \mathbb{R}^{L \times M} \), where \( M \) denotes the hidden dimension of the model. To effectively capture relevant information from the textual data across time, we employ a multi-head attention mechanism \cite{35}. A simple linear layer is then used to project \( S \) into a new dimension \( d_m \), resulting in \( S' \in \mathbb{R}^{L \times d_m} \). 

During this process, self-attention operations are applied. For each attention head \( k = \{1, \dots, K\} \), we define the query matrix \( Q_k \), the key matrix \( K_k \), and the value matrix \( V_k \) for the self-attention mechanism. Specifically, the query matrix is defined as \( Q_k = S' W^Q_k \), the key matrix as \( K_k = S' W^K_k \), and the value matrix as \( V_k = S' W^V_k \), where \( W^Q_k \), \( W^K_k \), and \( W^V_k \) are learnable weight matrices of shape \( \mathbb{R}^{d_m \times d} \). Here, \( d = d_k = \left\lfloor \frac{d_m}{K} \right\rfloor \) represents the dimension for each attention head. These matrices are then processed by the attention mechanism as follows:
{\footnotesize
\begin{equation}
Z_k = \text{ATTENTION}(Q_k, K_k, V_k) = \text{SOFTMAX}\left(\frac{Q_k K_k^\top}{\sqrt{d_k}}\right) V_k
\end{equation}
}


By utilizing this approach, the model effectively captures the temporal relationships within the textual data, improving its ability to understand semantic information and extract relevant features. The self-attention mechanism allows the model to focus on the most important time steps, enhancing prediction accuracy. After calculating the attention outputs for each head, the values \( Z_k \) are aggregated to form a final representation \( Z \in \mathbb{R}^{L \times d_m} \), where \( L \) is the sequence length and \( d_m \) is the feature dimension. This representation captures both temporal and semantic aspects of the data. 

Next, this representation is linearly projected into the hidden dimension \( D \) of the backbone model, producing a vector \( I \in \mathbb{R}^{L \times D} \). Finally, the text prompt embeddings are processed with the ReLU activation function, enabling the model to learn non-linear relationships and focus on the most important features.

\subsection{Time Series Input Embedding}
We preprocess the time series input using reversible instance normalization (RevIN) to mitigate distribution shifts in the data, as outlined in \cite{36}. This technique helps address the discrepancies in data distribution that often arise between the training and testing phases, ensuring the model generalizes effectively across various domains or time periods. By normalizing the data in a reversible manner, RevIN maintains the original characteristics of the data while aligning it to a more consistent distribution, which is crucial for stable model training.

Next, we partition the time series into multiple consecutive patches \cite{20}. Each patch has a fixed length \( L_p \), and these patches can be either overlapping or non-overlapping, depending on the stride \( S \). The total number of patches \( P \) is computed as:
\begin{equation}
P = \left\lfloor \frac{L - L_p}{S} \right\rfloor + 2
\end{equation}

This partitioning process divides the continuous time series into smaller, manageable segments, each containing a subset of the data. By splitting the time series into patches, the model can capture local temporal patterns while also being better equipped to handle long-term dependencies. These patches enable the model to focus on smaller, contextually relevant chunks of data, allowing it to capture nuanced contextual information at each time step and model the sustained impact of past events on future predictions.
Each patch \( \mathbf{X}_i \) represents the \( i \)-th patch is defined as:
\begin{equation}
\mathbf{X}_i = \{ x_t \ | \ t \in [(i-1) \cdot S, (i-1) \cdot S + L_p -1] \}, \forall i \in \{1, 2, \dots, P\}
\end{equation}

The final time series embedding is obtained by concatenating these patches:
\begin{equation}
\mathbf{X} = \begin{bmatrix} \mathbf{X}_1 \\ \mathbf{X}_2 \\ \vdots \\ \mathbf{X}_P \end{bmatrix} \in \mathbb{R}^{P \times D}
\end{equation}
where \( D \) is the feature dimension of the backbone language model.

\subsection{Frozen Pre-trained Model and Output Projection}
We refine the frozen backbone language model by fine-tuning the position embeddings and layer normalization layers. These adjustments help the model better handle sequential data and improve its adaptation to the task. After conducting a series of experiments, we decided to proceed with the initial approach, which showed the best performance.


After extracting the embeddings from the backbone model, we removed the dual-prompt prefix component to simplify the model. This adjustment enables us to concentrate solely on the learned embeddings derived from the input data. We then flattened the output embeddings into a one-dimensional vector for easier processing. Finally, a linear projection was applied to map these embeddings to the prediction space, resulting in the final output.

\section{Experiments} \label{sec-4}

\subsection{Datasets}


Recently, several public multimodal time series datasets have been developed. One notable dataset is a textual-numerical time series prediction dataset derived from the GDELT database, as presented in \cite{10}. This database catalogs global events alongside their related media reports, highlighting the significant impact of news on our daily lives.  For our experiments, we identified Nummentions as a critical variable for prediction since it relates to the attention given to specific event types within designated time frames and geographical areas. The dataset is categorized into ten distinct time root types, as illustrated in Table \ref{tab:1}. We gathered data from 55 regions across the United States, including national-level data. After cleaning and preprocessing the data, we filtered the dataset to include 53 regional datasets along with the national dataset for training and evaluation. This covers the period from August 17, 2022, to July 31, 2023, with daily frequency.

The work described in \cite{37} achieved fine-grained modality alignment by carefully selecting data sources and implementing strict filtering steps. This led to the introduction of the first multi-domain, multi-modal time series dataset, Time-MMD. After cleaning and preprocessing the data into a unified format, we chose datasets from two different domains for our study: agricultural data collected monthly and public health data from the United States collected weekly. This choice ensured comprehensive coverage for our experiments. From these datasets, we identified OT as the key target variable for prediction.

\begin{table}[hbtp]
\centering
{\footnotesize
\begin{tabular}{c|c}
\hline
\textbf{Event Number} & \textbf{Event Type Name} \\
\hline
01 & Make Public Statement \\
02 & Appeal \\
03 & Express Intent to Cooperate \\
04 & Consult \\
05 & Engage in Diplomatic Cooperation \\
07 & Provide Aid \\
08 & Yield \\
11 & Disapprove \\
17 & Coerce \\
19 & Fight \\
\hline
\end{tabular}
\caption{Event Types in GDELT Dataset}
\label{tab:1}
}
\end{table}

\begin{table*}[htbp]
\centering
\footnotesize
\setlength{\tabcolsep}{2pt}  
\begin{tabular}{c|ccccccccccc}
\hline
\textbf{Event} & \textbf{DLinear} & \textbf{NLinear} & \textbf{PatchTST} & \textbf{Autoformer} & \textbf{Informer} & \textbf{Transformer} & \textbf{iTransformer} & \textbf{TimeLLM} & \textbf{GPT4TS} & \textbf{GPT4MTS} & \textbf{Ours} \\
\hline
01 & 0.738 & 0.795 & 0.760 & 0.790 & 0.761 & 0.782 & 0.797 & 0.832 & 0.765 & \underline{0.717} & \textbf{0.709}\\
02 & 0.743 & 0.787 & 0.759 & 0.767 & \underline{0.727} & 0.815 & 0.779 & 0.816 & 0.756 & 0.735 & \textbf{0.719}\\
03 & 0.901 & 0.953 & 0.921 & 0.925 & 0.921 & 0.930 & 0.940 & 0.975 & 0.934 & \underline{0.896} & \textbf{0.855}\\
04 & 0.841 & 0.903 & 0.862 & 0.894 & 0.847 & 0.843 & 0.865 & 0.930 & 0.863 & \underline{0.831} & \textbf{0.819}\\
05 & \underline{0.984} & 1.096 & 1.079 & 1.027 & 1.005 & \textbf{0.951} & 1.173 & 1.091 & 1.046 & 1.028 & 0.985\\
07 & \underline{1.218} & 1.269 & 1.258 & 1.240 & 1.237 & 1.244 & 1.304 & 1.261 & 1.260 & 1.223 & \textbf{1.185}\\
08 & 0.971 & 1.000 & 0.999 & 0.989 & 0.977 & 1.017 & 1.017 & 1.024 & 0.989 & \underline{0.963} & \textbf{0.956} \\
11 & 1.102 & 1.164 & 1.110 & 1.113 & 1.151 & 1.070 & 1.171 & 1.181 & 1.083 & \underline{1.047} & \textbf{1.019}\\
17 & 0.941 & 1.005 & 0.966 & 1.019 & 0.996 & 0.971 & 1.026 & 1.029 & 0.952 & \underline{0.915} & \textbf{0.903}\\
19 & 1.698 & 1.691 & 1.652 & 1.639 & 1.618 & 1.683 & 1.743 & 1.685 & 1.630 & \textbf{1.612} & \underline{1.616}\\

\hline

Average & 1.013 & 1.066 & 1.036 & 1.040 & 1.024 & 1.031 & 1.082 & 1.082 & 1.028 & \underline{0.997} & \textbf{0.976} \\

\hline
\end{tabular}
\caption{Comparison results for the GDELT dataset, grouped by event and MSE. The lower the better. Bold indicates the best result, while underlined indicates the second best.}
\label{tab:2}
\end{table*}

\begin{table*}[htbp]
\centering
\small  
\setlength{\tabcolsep}{2pt}  
\begin{tabular}{c|ccccccccccc}
\hline
\textbf{Event} & \textbf{DLinear} & \textbf{NLinear} & \textbf{PatchTST} & \textbf{Autoformer} & \textbf{Informer} & \textbf{Transformer} & \textbf{iTransformer} & \textbf{TimeLLM} & \textbf{GPT4TS} & \textbf{GPT4MTS} & \textbf{Ours} \\
\hline
01 & 0.639 & 0.686 & 0.661 & 0.683 & 0.656 & 0.667 & 0.677 & 0.717 & 0.660 & \underline{0.639} & \textbf{0.633} \\
02 & 0.652 & 0.685 & 0.663 & 0.678 & 0.653 & 0.688 & 0.675 & 0.707 & 0.660 & \underline{0.650} & \textbf{0.640} \\
03 & \underline{0.718} & 0.753 & 0.731 & 0.740 & 0.730 & 0.736 & 0.739 & 0.775 & 0.735 & 0.722 & \textbf{0.701} \\
04 & 0.693 & 0.735 & 0.707 & 0.732 & 0.700 & 0.693 & 0.711 & 0.756 & 0.707 & \underline{0.691} & \textbf{0.683} \\
05 & \underline{0.760} & 0.829 & 0.815 & 0.809 & 0.779 & \textbf{0.750} & 0.846 & 0.831 & 0.802 & 0.796 & 0.773 \\
07 & \underline{0.799} & 0.841 & 0.832 & 0.830 & 0.815 & 0.813 & 0.841 & 0.839 & 0.830 & 0.817 & \textbf{0.793} \\
08 & \underline{0.732} & 0.760 & 0.759 & 0.765 & 0.746 & 0.763 & 0.766 & 0.770 & 0.756 & 0.742 & \textbf{0.731} \\
11 & 0.755 & 0.798 & 0.770 & 0.785 & 0.783 & 0.750 & 0.794 & 0.815 & 0.762 & \underline{0.747} & \textbf{0.729} \\
17 & 0.730 & 0.772 & 0.754 & 0.781 & 0.757 & 0.749 & 0.770 & 0.792 & 0.748 & \underline{0.729} & \textbf{0.721} \\
19 & 0.789 & 0.819 & 0.806 & 0.843 & 0.795 & 0.837 & 0.832 & 0.810 & 0.790 & \textbf{0.770} & \textbf{0.770} \\
\hline
Average & \underline{0.727} & 0.768 & 0.750 & 0.765 & 0.741 & 0.744 & 0.765 & 0.781 & 0.745 &0.730 & \textbf{0.717} \\		 	 	 
\hline
\end{tabular}
\caption{Comparison results for the GDELT dataset grouped by event and MAE. The lower the better. Bold indicates the best result, while underlined indicate the second best.}
\label{tab:3}
\end{table*}

\subsection{Baselines and Experimental Settings}
We selected several state-of-the-art (SOTA) methods for time series forecasting as baselines. These methods include a variety of advanced models, such as Transformer-based architectures like PatchTST \cite{20}, Autoformer \cite{38}, Informer \cite{26}, and iTransformer \cite{25}, along with two linear models: DLinear and NLinear \cite{9}. Furthermore, we incorporated three approaches based on pre-trained language models: GPT4TS \cite{15}, TimeLLM \cite{16}, and GPT4MTS \cite{10}, all of which utilize the same GPT-2 base backbone language model uses 6 layers.  

All models were evaluated under a consistent experimental setup to ensure a fair comparison. In this setup, we divided the dataset into training, validation, and test sets with a ratio of 7:2:1. For the GDELT dataset, which operates on a daily time scale, we implemented a unified lookback window size of 15 to predict a future duration of $T=7$ days follow \cite{10}. Similarly, for the public health (US) dataset, which has a weekly time frame, and the agriculture dataset, on a monthly basis, we applied lookback window sizes of 36 and 12, respectively, to forecast future spans of 12 weeks and 4 months.

Both the baselines and DP-GPT4MTS were implemented using PyTorch and run on a server equipped with multiple 32GB Tesla V100 GPUs. Hyperparameters were carefully tuned based on their performance on the validation set. We initially set the learning rate to 0.001 and established a maximum of 20 training epochs, with training ceasing when the validation loss showed no improvement for three consecutive epochs. For each experiment, we performed three independent runs using different random seeds, and the results were averaged to ensure the stability and reliability of the performance metrics.

\subsection{Performance Comparison}
We utilize the commonly used techniques of Mean Squared Error (MSE) \cite{40} and Mean Absolute Error (MAE) \cite{41} as evaluation metrics to assess the performance of all methods. 

The results of the time series predictions for the GDELT dataset, which has a daily frequency, are presented in Tables \ref{tab:2} and \ref{tab:3}. Furthermore, Table \ref{tab:4} showcases the prediction results for the two selected datasets from Time-MMD, i.e., the Public Health (US) dataset with a weekly frequency, and the Agricultural sector dataset with a monthly frequency. In these tables, values marked in \textbf{bold} and \underline{underlined} signify the best and second-best performances, respectively, emphasizing the relative improvement of the DP-GPT4MTS model over the top baseline models.


Our analysis of the experimental results shows that the proposed model consistently outperforms baseline approaches across various textual-numerical datasets, regardless of the domain or temporal resolution of the time series data. For instance, our model achieves the lowest MSE on 8 out of 10 events in the GDELT datasets, and it achieves the lowest MAE on 9 out of 10 events. Additionally, it secures the best average MSE and MAE overall in the GDELT datasets. For the Time-MMD dataset, the MSE and MAE of our model for the two domains, Agriculture and Public Health, are 0.098, 0.211, 0.890, and 0.601, respectively. Each of these is the best among all the models. This outcome highlights the significant benefits of incorporating textual information into time series forecasting through our dual-prompt mechanism, which enhances prediction accuracy.



It should be noted that the GPT4TS and GPT4MTS models showed suboptimal performance on the Agriculture and Public Health datasets. This may be due to the presence of redundant or irrelevant information in the Time-MMD dataset. Specifically, the inclusion of "Not Available" (NA) markers and unnecessary textual elements likely creates noise that negatively impacts the reasoning capabilities of LLMs. In contrast, our DP-GPT4MTS model demonstrates remarkable resilience, maintaining high-performance levels despite these challenges. This robustness showcases its ability to filter out extraneous information and focus on the most critical features, ensuring accurate and reliable predictions. Overall, these findings confirm the effectiveness of our approach in tackling the complexities involved in textual-numerical time series forecasting tasks.

\begin{table}[ht]
\centering
\small 
\setlength{\tabcolsep}{9pt}  
\begin{tabular}{cccccc}
\hline
\multirow{2}{*}{\textbf{Models}} & \multicolumn{2}{c}{\textbf{Agriculture}} & \multicolumn{2}{c}{\textbf{Public Health}} \\
\cline{2-5}
 & \textbf{MSE} & \textbf{MAE} & \textbf{MSE} & \textbf{MAE} \\
\hline
\textbf{DLinear} & 0.411  & 0.440 & 1.465 & 0.834 \\   	 	 	
\textbf{Nlinear} & 0.116  & 0.244 & 1.126 & 0.722 \\ 
\textbf{PatchTST} & 0.105 & \underline{0.213} & \underline{1.020} & \underline{0.658} \\
\textbf{Autoformer} & 0.271 & 0.373 & 2.196 & 1.197 \\
\textbf{Informer} & 0.165 & 0.275 & 2.165 & 0.985 \\
\textbf{Transformer} & 2.505 & 1.128 & 1.233 & 0.731 \\
\textbf{iTransformer} & 0.124 & 0.243 & 1.116 & 0.677 \\
\textbf{TimeLLM} & 0.198 & 0.245 & 1.060 & 0.685 \\
\textbf{GPT4TS} & \underline{0.103} & 0.217 & 1.082 & 0.681 \\
\textbf{GPT4MTS} & 0.106 & 0.214 & 1.033 & 0.668 \\
\textbf{Ours} & \textbf{0.098} & \textbf{0.211} & \textbf{0.890} & \textbf{0.601} \\
\hline
\end{tabular}

\caption{Comparison results for the Time-MMD datasets using MSE and MAE metrics. The lower the better. Bold indicates the best result, while underlined indicates the second best.}
\label{tab:4}
\end{table}

\begin{table}[ht]
\centering
\small  
\setlength{\tabcolsep}{3pt}  
\begin{tabular}{c|ccccc}
\toprule
\multirow{2}{*}{\textbf{Event}} & \multicolumn{4}{c}{\textbf{Variant}}  & \multirow{2}{*}{\textbf{DP-GPT4MTS}}   \\
\cline{2-5}
 & \textbf{SEP} & \textbf{STP} & \textbf{DP-NTSA} & \textbf{SPET} \\
\midrule
1 & 0.719 & 0.766 & 0.740 & 0.727 & \textbf{0.709} \\
2 & 0.723 & 0.773 & 0.737 & 0.732 & \textbf{0.719}\\
3 & 0.891 & 0.929 & 0.910 & 0.907 & \textbf{0.855} \\
4 & \textbf{0.815} & 0.866 & 0.839 & 0.828 & 0.819 \\
5 & 1.049 & 1.091 & 1.035 & 1.059 & \textbf{0.985} \\
7 & 1.218 & 1.260 & 1.244 & 1.234 & \textbf{1.185} \\
8 & 0.967 & 1.014 & 0.982 & 0.974 & \textbf{0.956} \\
11 & 1.037 & 1.107 & 1.065 & 1.043 & \textbf{1.019}\\
17 & 0.924 & 0.971 & 0.932 & 0.915 & \textbf{0.903}\\
19 & 1.625 & 1.626 & 1.640 & 1.621 & \textbf{1.616} \\
\midrule
\textbf{Average} & 0.997 & 1.040 & 1.012& 1.004 & \textbf{0.976} \\
\bottomrule
\end{tabular}
\caption{Ablation study results for different variants on the GDELT dataset with MSE metric.}
\label{tab:5}
\end{table}

\subsection{Model Analyse}

\textbf{Ablation Study} To evaluate the effectiveness of two key innovations of our model including the dual-prompt mechanism and the event self-attention operation, we conducted an ablation study using the GDELT dataset. We kept the hyperparameters consistent with the original settings and established the following variants of DP-GPT4MTS:

\begin{enumerate} 

\vspace{-0.5ex}
\item[(1)] Single Explicit Prompt (SEP): This variant uses only the explicit prompt to guide the reasoning of the pre-trained language model while ignoring the textual information that accompanies the numerical data.

\vspace{-0.5ex}

\item[(2)] Single Textual Prompt (STP): In this variant, we rely solely on the textual prompt, embedding and training the textual information as a soft prompt to direct the large language model.

\vspace{-0.5ex}

\item[(3)] Dual-Prompt Mechanism without Textual Embedding Self-Attention (DP-NTSA): This approach eliminates the self-attention mechanism during the training of the textual embedding prompt within the dual-prompt framework.

\vspace{-0.5ex}

\item[(4)] Swapped Positions of Explicit and Textual Prompts (SPET): In this variation, we change the arrangement of the explicit and textual prompts by swapping their positions within the dual-prompt mechanism.
\end{enumerate}


The findings presented in Table \ref{tab:5} highlight the effectiveness of the key components in DP-GPT4MTS. Notably, incorporating textual information through the dual prompt mechanism significantly improves the model's performance. The comparison of the SEP variant with DP-GPT4MTS indicates that the textual context accompanying the numerical data plays a crucial role in guiding the model's reasoning process.  The results of STP reinforce the necessity for the model to process both structured numerical data and unstructured textual information cohesively. Furthermore, the results of DP-NTSA illustrate the importance of the self-attention mechanism, which allows the model to extract the most relevant information effectively. The findings from SPET suggest that utilizing explicit hard prompts provides the model with clear direction, emphasizing that hard prompt embeddings should precede soft textual prompts.  Overall, these findings underscore how prompt design and information flow impact the model's predictive accuracy.


\begin{figure}[hbtp] 
    \centering
    \includegraphics[width=0.38\textwidth]{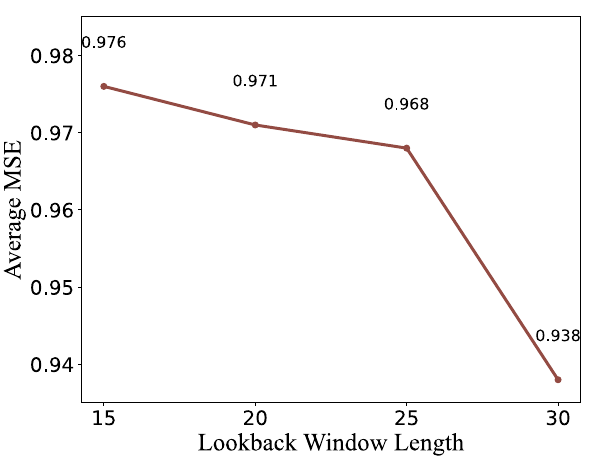} 
    \caption{Results of Hyperparameter Study} 
    \label{fig:6} 
\end{figure}

\noindent\textbf{Hyperparameter study} We evaluated the model's performance on textual-numerical time series data with different lookback window sizes. Figure \ref{fig:6} shows the experimental results conducted on the GDLET dataset. These results indicate that as the lookback window size increases, the prediction error (average MSE) decreases. A larger lookback window allows the model to capture more historical information, thereby improving its ability to recognize long-term dependencies in textual-numerical time series. Furthermore, a larger lookback window can reduce the influence of noise in the data, improving prediction accuracy and further validating the model's generalization ability across different scenarios. However, a huge window may lead to an increase in computational complexity, thereby reducing the operational efficiency of the model. Therefore, selecting an appropriate lookback window length is important for time series forecasting. To avoid increasing the computational complexity, we choose 15 as the default value of the lookback window length in the previous experiments, for which the results and analysis suggest that the model outperforms existing models on diverse contextual-numerical time series datasets.


\section{Conclusion and Future work} \label{sec-5}

This paper presents a dual-prompt large language model framework for processing textual-numerical time series data, where each timestamp is associated with contextual text. Leveraging a self-attention mechanism, the framework generates high-quality embeddings that capture both global and salient information. Extensive experiments across diverse domains and temporal granularities demonstrate the model’s strong performance and robustness.
Unlike traditional single-prompt models, our approach integrates soft prompts to guide contextual understanding and hard prompts to emphasize key information, enabling more precise and informative text representations. This dual-prompt design significantly enhances inference quality and prediction accuracy by fully exploiting textual context. Overall, our method offers a flexible and effective solution for complex textual-numerical time series modeling, and opens new avenues for large language model applications in this area.

We observed that single-prompt large language models underperformed compared to univariate time series methods on certain datasets. This may be attributed to the excessive redundancy and noise present in textual information, which can obscure key details. Specifically, when processing texts related to news and policy events, the structure and quality of the text significantly impact model performance. Lengthy descriptions, irrelevant background information, and potential emotional biases can introduce noise, hindering the model's reasoning and prediction capabilities. This highlights the challenge of balancing numerical and textual information in textual-numerical time series datasets to ensure the relevance and quality of textual data. 


For future work, we recommend prioritizing the construction of more comprehensive and high-quality textual-numerical time series benchmark datasets that encompass real-world data across multiple domains and time periods. This will provide a robust foundation for the application and validation of large language models. In addition to dataset development, exploring zero-shot and few-shot learning methods based on these datasets can unlock the potential of large models in scenarios with limited samples. These studies will not only advance textual-numerical time series prediction techniques but also offer new insights and directions for broader artificial intelligence applications.



\bibliography{aaai2026}
\end{document}